# HOG, LBP and SVM based Traffic Density Estimation at Intersection

Devashish Prasad, Kshitij Kapadni, Ayan Gadpal, Manish Visave, Kavita Sultanpure
Pune Institute of Computer Technology, India
{devashishkprasad,kshitij.kapadni,ayangadpal2,manishvisave149,kavita.sultanpure}@gmail.com

*Abstract*— **Increased amount of vehicular traffic on roads is a significant issue. High amount of vehicular traffic creates traffic congestion, unwanted delays, pollution, money loss, health issues, accidents, emergency vehicle passage and traffic violations that ends up in the decline in productivity. In peak hours, the issues become even worse. Traditional traffic management and control systems fail to tackle this problem. Currently, the traffic lights at intersections aren't adaptive and have fixed time delays. There's a necessity of an optimized and sensible control system which would enhance the efficiency of traffic flow. Smart traffic systems perform estimation of traffic density and create the traffic lights modification consistent with the quantity of traffic. We tend to propose an efficient way to estimate the traffic density on intersection using image processing and machine learning techniques in real time. The proposed methodology takes pictures of traffic at junction to estimate the traffic density. We use Histogram of Oriented Gradients (HOG), Local Binary Patterns (LBP) and Support Vector Machine (SVM) based approach for traffic density estimation. The strategy is computationally inexpensive and might run efficiently on raspberry pi board. Code is released at https://github.com/DevashishPrasad/Smart-Traffic-Junction.**

*Keywords—Image processing, Computer vision, Traffic estimation, Traffic density, HOG, Histogram of Oriented Gradients, LBP, Local Binary Patterns, Smart traffic junction, Traffic intersection, Intelligent traffic control system, Machine learning, SVM, Support Vector Machine.*

## I. INTRODUCTION

The signal transition timing slots in existing traffic light systems are static and that they have fixed time cycles. At the intersection, even though a specific lane has less traffic, the lane is allotted with a fixed time which ends in inefficient management of traffic. Though the road traffic authorities attempt to mitigate the matter by switching different time for traffic light on the basis of the time of day, this is often ineffective as traffic is usually random for the larger period of time. Inefficient management of traffic ends up in several problems; the vehicles get trapped within the redundant traffic congestion. Some times within the normal conditions they need to wait for a longer time-span although the traffic density is extremely less in other or all lanes. Other issues that a traditional traffic light system contributes are pollution, money loss, health issues, accidents, emergency vehicle passage and traffic violations that ends up in the decline in productivity. This demands sensible and smart traffic light systems that are adaptable to traffic density and modify the traffic light timings accordingly.

## II. LITERATURE REVIEWS

There are several techniques projected for traffic density estimation. This additionally includes sensors and other hardware based solutions. Image processing is most preferred as a result of its cost effective than hardware based solutions. Image processing based traffic density estimation techniques solely need cameras that are pre-installed on most of the junctions and it demands no other sort of sensors. There are several image processing techniques for traffic density estimation like background subtraction, motion detection, edge detection, HOG, LBP and deep learning neural networks based strategies. Each of these strategies has its own benefits and drawbacks.

Background subtraction based density estimation strategies proposed by SS. Shi [1] take a reference image of road with zero traffic. Then the frame subtraction is performed with subsequent frames scan from camera in real time having traffic. The subtraction leads to generation of frame having pixels of solely traffic vehicles. This is also often referred to as foreground extraction. During this method, the density of traffic is calculated by evaluating these traffic vehicle pixels. The density is more if more pixels of traffic are found.

There are some enhancements in background subtraction techniques as Liyuan Li [2] proposed a Bayesian framework that incorporated spectral, spatial, and temporal features to characterize the background appearance. Ying Ren [3] proposed a background extraction methodology that concerned calculating the mean of the background Gaussian distribution within the background map.

M. S. Uddin, [4] used edge detection technique for traffic density estimation. The amount of pixel of edges were calculated and compared to seek out the traffic density estimation. J. Vijayaraj [5] performed a comparative study of varied image process techniques used for traffic congestion management of vehicles based on edge detection as well as background subtraction. Kavya P Walad [6] performed traffic light control system using edge detection techniques.

Yeshwanth [7] used deep neural network based approach to seek out traffic density estimation. Convolutional neural network (CNN) was used for traffic vehicles detection and density estimation. A comparative study of CNN and background subtraction was conducted to work out accuracy results. CNN performed more accurate than background subtraction strategies.

Background subtraction strategies are liable to the amendment in environment. If the reference image is taken within the day-light, then as the day-light gets low, the subtraction ends up in terribly poor estimations. Deep learning based strategies have a decent accuracy however needs a lot of training data and moreover have high computational requirements. Edge detection based strategies calculate range of pixels out of edges but in real time there may be edges of trees, people, divider and light illuminations which may lead to false density estimations.

We use HOG and LBP feature extraction with SVM based algorithm for traffic density estimation. The HOG descriptor was introduced by Dalal and Triggs [8]. During this paper, the calculation of HOG functions relies on edge gradient orientation histograms for car detection. This system counts occurrences of gradient orientation in localized parts of a picture. Native object look and form will usually be characterized very well by the distribution of native intensity gradients or edge direction.

A. Mukhtar and T. B. Tang [9] proposed a vision based technique for motorcycle detection using the HOG features and SVM for classification. Z. Chen, K. Chen and J. Chen, [10] proposed a system for Vehicle and Pedestrian Detection Using Support Vector Machine and Histogram of Oriented Gradients Features. L. Weixing, S. Haijun, P. Feng, G. Qi and Q. Bin, [11] proposed fast pedestrian detection via modified HOG feature and reducing the number of features produced by the HOG descriptor.

LBP is the particular case of the Texture Spectrum model proposed in 1990 [12][13]. LBP was first described in 1994[14]. It was found to be a powerful feature for texture classification and it has further been determined that when combining LBP with the Histogram of oriented gradients (HOG) descriptor improves the detection performance. We use HOG and LBP features with SVM for traffic density estimation which is quite effective and therefore the paper elaborates an extension to create these strategies more practical.

III. METHODOLOGY

A. Proposed method

In the proposed method, we first train the SVM. For this, we first create a data set from the QMUL junction 2 video[15]. We assume that it was captured from a CCTV of one lane of an intersection or junction. From this video we save 200 frames as training images. For each training image we extract the Region of Interest (ROI) which consists of the region of image which we want to focus on. We select ROI such that only traffic is included and all the unnecessary portions not to be focused in the original frame are excluded. Then the ROI is divided into a grid of 8X9 cells with each cell of 44x44 pixels. The number of cells and the size of grid may vary based on the ROI which in turn varies from lane to lane and intersection to intersection. All of these cells are saved for each frame. Then for each cell we extract the HOG and LBP features and train the SVM for binary classification that the cell has traffic or road. The trained model is serialized and saved to the disk.

For traffic density estimation, we take the real time feed from the cameras and extract the ROI and again create the grid of 8x9 cells with each cell of 44x44 pixels. Then we load the saved SVM model and pass each of these cells to it for prediction. The SVM gives the prediction for each of these cells that it contains road or traffic. In this way, we get a binary classification for whole grid and from this we can get estimation of traffic density by getting the number of cells having traffic.

B. Architecture

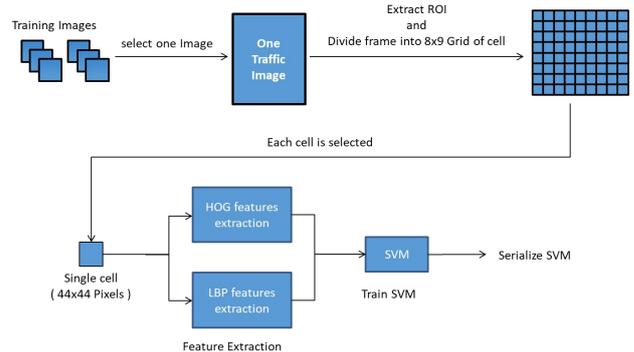

Fig. 1 Flow of execution for training SVM

For training the SVM we first create the training images from a QMUL junction 2 video. For each of these images we extract ROI. Then we divide the image into grid of cells of 8X9. Each cell is of 44X44 pixels. A cell may contain traffic or road. For each of these cells we extract the HOG features and LBP features. We make a feature vector by concatenating both features. We extract features from all cells of all images. Then we train the SVM on these features for binary classification. The trained SVM will able to predict that a cell contains traffic vehicle or road. Fig 1. shows the steps in flow of training the SVM.

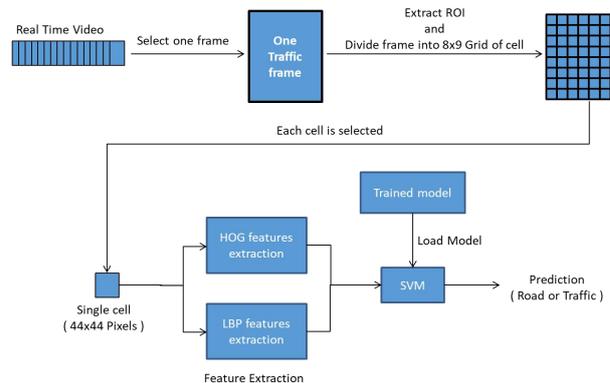

Fig. 2 Flow of execution for prediction

The steps shown in Fig. 2 are carried for traffic density estimation on a lane of intersection. We use camera installed on a particular side of the intersection to get the frames of that lane. We extract frames from the video and extract the same ROI as that of training images. Again we divide the image into grid of cells of 8X9 pixels. We again extract HOG and LBP features from each of these cells and make a feature vector. We then load our pre-trained SVM model and pass it this feature vector. SVM gives the prediction for each cell that it contains traffic or road. We then get number of cells that contain only traffic and number of cells that contains no traffic (only road). We use this information to estimate traffic density.

## IV. SYSTEM IMPLEMENTATION

### A. Reading Image

The real time image is read from the video or camera in the RGB format. Fig. 3 shows the image read from the the QMUL junction 2 video.

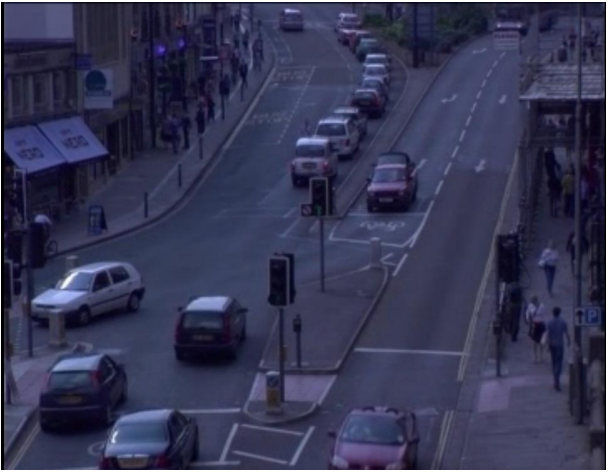

Fig. 3. Image read from camera

### B. Extract ROI

As the position of camera is fixed we can directly extract Region of Interest from the whole frame. We ignore the unwanted portions of the frame by clipping them off from ROI. The unwanted portions in the frame may contain trees, footpaths and other irrelevant objects. Extracting ROI helps us reducing unnecessary calculations and improves the accuracy of density estimation by minimizing the errors. Fig. 4 shows the Extracted Region of Interest from original image.

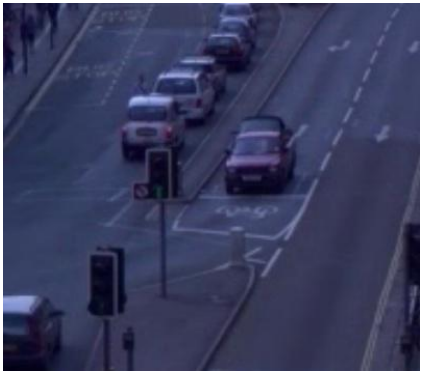

Fig. 4. Extracted Region of Interest

### C. Grid cells

We divide the extracted ROI into the grid with cells of 8X9 as shown in the Fig. 5 we divide the frame in grid of cells because we want to identify the portions of image which contains traffic and portions of image which don't contain traffic. The size of single cell needs to be tuned according to kind of image we get from CCTV camera. And it totally depends on the position of CCTV cameras. Dividing the image in grid also helps in calculations of HOG and LBP feature extraction.

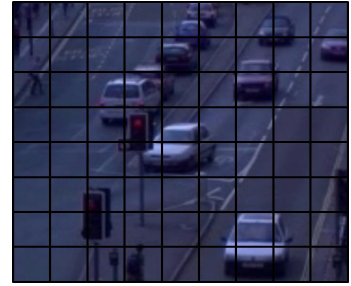

Fig. 5 ROI with grid of cells

### D. Extracting HOG features

The HOG features are extracted from the each cells made on the image ROI. Fig. 6 shows the computed HOG image for one cell. For each cell we again divide it further into more sub cells for calculating its HOG. We divide each cell into sub cell of 3X3 pixel cell and compute HOG over 8 orientations. We get final HOG feature vector of 1568 dimensions.

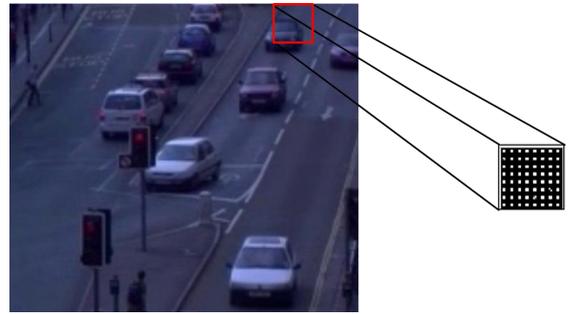

Fig. 6 Extracted HOG image of one cell

### E. Extracting LBP features

The LBP features are extracted from the same cells made on the image ROI. LBP requires gray scale image as an input. We define two parameters for LBP, the radius and the number of points around outer radius whose value depend on various factors like size of a cell. Fig. 7 shows the computed LBP image for one cell. We generate histogram out of the LBP image of the cell and normalize it. The final LBP feature vector extracted is 26 dimensional.

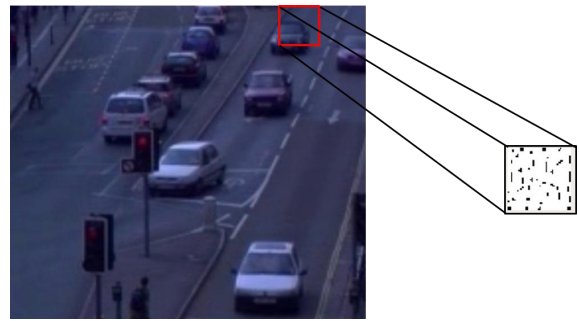

Fig. 7 Extracted LBP image of one cell

### F. Training SVM

We first train the SVM on the training images. For creating training dataset we first create the training images from the QMUL junction 2 video. We save frames of video as training images. For each of these images we extract ROIs as stated earlier. Then we divide the image into grid of cells of 8X9 as stated earlier where each cell is of 44X44 pixels. And a cell may contain traffic or road. For each of these cells we extract the HOG features and LBP features as explained earlier. We make a feature vector by concatenating both feature vectors. So, we get our final 1594 (1568+26) dimensional feature vector. Then we train the SVM on these features for binary classification between cells having traffic and cells having road.

We saved total 200 images from the QMUL junction 2 video and extracted 20000 cells. The cells were then divided into two classes as traffic and not traffic. The model was evaluated using 80-20 data set split.

Accuracy states how often the model is correct. Accuracy is (truepositives+truenegatives)/totalexamples. Accuracy score of the SVM model on test set was calculated to be 0.9488248. Precision is when model predicts positive, how often it is correct. Precision was calculated to 0.9400322. The Recall helps when the cost of false negative is high. It was calculated to 0.9402985. And the F1 Score combines the precision and recall and was calculated to 0.9398987.

### G. Traffic Density Estimation

Z. Zhiqian Chen [10] performed Vehicle and Pedestrian Detection Using SVM and HOG Features. The result performed good in both accuracy and speed of the developed system, However the HOG is good only at capturing edges and corners in images. LBP captures the local patterns. Ultimately HOG and LBP captures different kinds of information, which make them complimentary to each other. Ashwin Arunmozhi [16] did the Comparison of HOG, LBP and Haar-Like Features for On-Road Vehicle Detection.

We use camera installed on a particular side of the intersection to get the frames of that lane or a video. We extract frames from the video and extract the same ROI as that of training images. We assume a cell contains traffic if it has 30% or more portion of traffic vehicle as shown in figure 8. We also put traffic signal lights in traffic class.

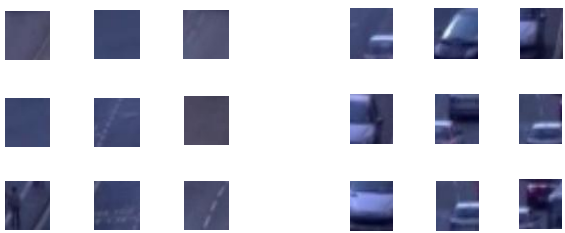

Fig. 8. Cells without traffic      Fig. 9. Cells with traffic

We again divide the image into grid of cells of 8X9 pixels. We again extract HOG and LBP features from each of these cells and make the feature vectors for each cell. We then load our pre-trained SVM model and pass these feature vectors for classification. The SVM classification is performed on each of the cells and then we can calculate the density of traffic by counting the number of cells having traffic. If the count of cells having traffic is more, then the density of traffic is more and vice versa. By doing this we get the accurate estimation of present traffic on that particular lane.

The following figures show results obtained using the above explained approach. We drew the rectangles representing the cells and coloured them according to the classification results of SVM for them respectively. We colour the cell red if it has traffic and green if it has road.

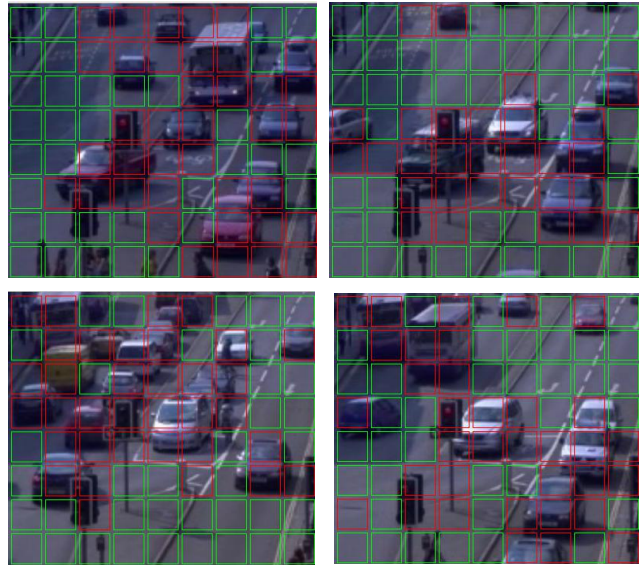

Fig. 10. Results on video frames

### CONCLUSION

In this paper, we analyzed various possible image processing techniques for traffic density estimation. The cameras installed on the intersection can be utilized to estimate the traffic density on each side. We found that the combination of HOG and LBP descriptor along with the Support Vector Machine (SVM) gave best results. This method can serve as a core part in intelligent traffic management system to control the traffic signals.